\pdfoutput=1

\documentclass[11pt]{article}

\usepackage{acl}

\usepackage{times}
\usepackage{latexsym}
\usepackage{graphicx}

\usepackage[T1]{fontenc}

\usepackage[utf8]{inputenc}

\usepackage{microtype}

\usepackage{inconsolata}

\title{Predicting Anti-microbial Resistance using Large Language Models}

\author{Hyunwoo Yoo, Bahrad Sokhansanj, James R. Brown, Gail Rosen \\
        Drexel University \\  \{hty23, bas44, jb4633, glr26\}@drexel.edu}

\begin{document}
\maketitle

\begin{abstract}
During times of increasing antibiotic resistance and the spread of infectious diseases like COVID-19, it is important to classify genes related to antibiotic resistance. As natural language processing has advanced with transformer-based language models, many language models that learn characteristics of nucleotide sequences have also emerged. These models show good performance in classifying various features of nucleotide sequences. When classifying nucleotide sequences, not only the sequence itself, but also various background knowledge is utilized. In this study, we use not only a nucleotide sequence-based language model but also a text language model based on PubMed articles to reflect more biological background knowledge in the model. We propose a method to fine-tune the nucleotide sequence language model and the text language model based on various databases of antibiotic resistance genes. We also propose an LLM-based augmentation technique to supplement the data and an ensemble method to effectively combine the two models. We also propose a benchmark for evaluating the model. Our method achieved better performance than the nucleotide sequence language model in the drug resistance class prediction.
\end{abstract}

\begin{figure}
    \centering
    \includegraphics[width=1\linewidth]{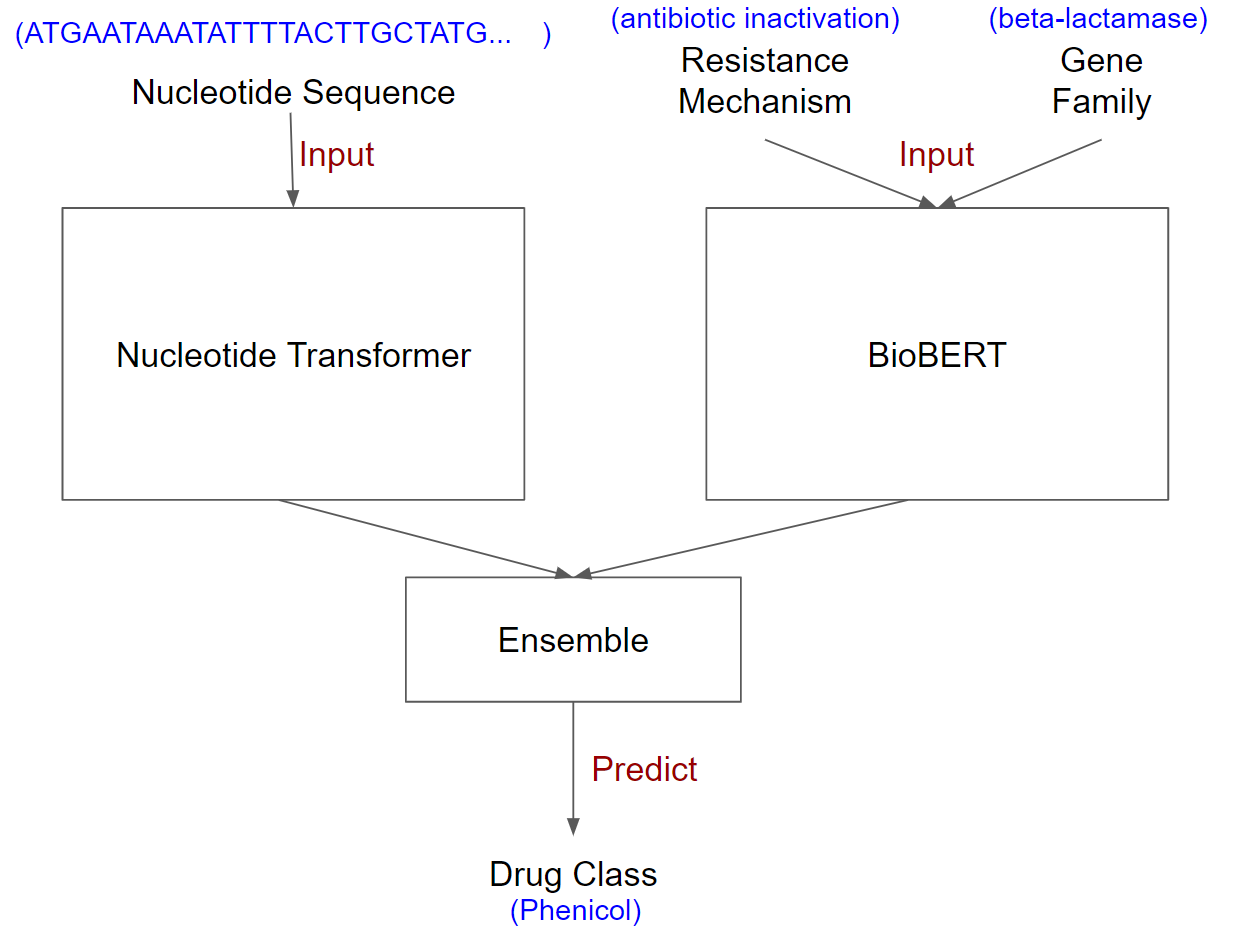}
    \caption{Overview of our approach}
\end{figure}

\begin{table*}
\centering
{\scriptsize 
\begin{tabular}{lll}
\hline
\textbf{Output} & \textbf{Input Example} & \textbf{BioBERT}\\
\hline
Base & Gene Family: Beta-lactamases, Resistance Mechanism: Antibiotic incativation &  78.20\\
Entity marker (punct) & [Gene Family]: Beta-lactamases, [Resistance Mechanism]: Antibiotic incativation & 77.41\\
Typed entity marker & *Beta-lactamases*, \#Resistance Mechanism\# &  77.70\\
Typed entity marker (punct) & *[Gene Family]: Beta-lactamases*, \#[Resistance Mechanism]: Antibiotic incativation\# &  78.46\\
\hline
\end{tabular}
}
\caption{\label{citation-guide}
 Test macro F1 score of different entity representation techniques in Antibiotic Resistance Drug Class Prediction with BioBERT.
}
\end{table*}

\section{Introduction}

The genes for antibiotic resistance have increased rapidly over the past 10 years and have become a threat to human health \cite{Zhang2022GlobalHealthRisk:1}. Moreover, dangerous infectious diseases like COVID-19 can also spread. In such times, it is important to classify the DNA sequences of antibiotic resistance genes. In bioinformatics, the main method for classifying DNA sequences has been to find similar sequences by aligning two DNA sequences using text alignment \cite{Bonin2023MEGAResAMR}. Recently, there have been methods that use language models created from the nucleotide or protein sequences of various species and fine-tune them to create classifiers\cite{Brandes2022ProteinBERT, Ji2021DNABERT, zhou2023dnabert2}. These methods have the advantage of being able to identify which parts of the nucleotide sequence are important. To fine-tune, databases containing information on antibiotic resistance genes must be used. The main databases are CARD \cite{Jia2017CARD} and MEGARes \cite{Doster2020}. Existing methods use the labels associated with antibiotic resistance genes, such as the class to which the resistance gene belongs, for example, the label of the antibiotic to which resistance is present. It is a prediction of a single label from a single gene sequence \cite{Kang2022FineTuningBERT}. However, if we look at the CARD or MEGARes databases, there are several attributes that describe a particular gene. There are Gene Family and Resistance Mechanism. If we use this information when predicting the antibiotic to which resistance is present, it could be helpful for prediction. Here, we get an idea and propose a model that uses human-readable information to predict antibiotic resistance genes. We also provide a method to merge the different classification systems of CARD and MEGARes. We will also explain the LLM-based data augmentation technique for rare classes with few samples.\\


\section{Approaches}

Our approaches include fine-tuning a pre-trained language model with various species' gene nucleotide sequence data to predict antibiotic resistance genes and their classes. We also fine-tune a pre-trained language model trained on a corpus containing diverse papers from the fields of biology and medicine to predict the names of antibiotic resistance gene properties. We provide an effective ensemble model \cite{Kumari2021} using the above two models in a weighted soft voting method. To integrate the classes, we combine the DNA sequences and the concepts that describe them from CARD and MEGARes into one. We use the EBI ARO ontology \cite{cook2016european} to combine CARD tagging and MEGARes tagging into one class system. For rare classes with few samples, we use BioGPT \cite{luo2022biogpt} prompting to perform data augmentation.

\subsection{Nucleotide Sequence Based Antibiotic Resistance Drug Class Classification}

Following the structure of \cite{dallatorre2023nucleotide}, we uses a large pre-training language model based on nucleotide sequences and fine-tune a classifier based on Drug Class data. The nucleotide sequence input is limited to a length of 1000, the input size of the pre-training model. The tokenizer uses a 6-mer tokenizer. A 6-mer tokenizer is a type of k-mer tokenizer. A k-mer tokenizer is a technique used in genome analysis and bioinformatics research that splits a biological sequence into substrings of length k \cite{mejiaguerra2019kmer}. The pre-training model uses NT, which is pre-trained on multi-species including bacteria, fungi, inverterbate, protozoa, verterbate gene sequences. Unlike other nucleotide sequence-based pre-training models that mostly use human genes, this model is trained on multi-species genes, providing a better representation. Fine-tuning is done using LoRA tuning. LoRA tuning is a method that fixes the weights of a pre-trained large-scale language model and inserts a low-rank decomposed matrix into each transformer layer, dramatically reducing the number of trainable parameters for the downstream task \cite{hu2021lora}. This allows for more effective fine-tuning.

\subsection{Text Information Based Antibiotic Resistance Drug Class Classification}

Text information based antibiotic resistance drug class classification uses a BioBERT language model pre-trained on a large medical and biological text corpus as the pre-training model. BioBERT is a pre-trained biomedical language representation model that uses a large-scale biomedical text corpus including PubMed abstracts, PMC full-text articles, and the Genia corpus. \cite{lee2020biobert} We fine-tune this model to extract antibiotic resistance drug classes, such as Drug Class or Gene Family, from text that describes antibiotic resistance genes. We aim to improve the performance of the classifier by utilizing a pre-trained biomedical text-based model. Instead of using multiple classification layers, we create a single classification layer and fine-tune it. The training data is structured as [Resistance Mechanism] followed by a description of the attribute, such as Antibiotic inactivation. To further improve performance, we create a format that encloses special characters \cite{zhou2021improved}, such as *[Gene Family]: Beta-lactamases*, \#[Resistance Mechanism]: Antibiotic inactivation\#.

\begin{table*}[h!]
\centering

\begin{tabular}{lllll}
\hline
\textbf{Method} & \textbf{Accuracy} & \textbf{Macro F1} & \textbf{Precision} & \textbf{Recall}\\
\hline
NT & 84.15 & 64.04 & 72.78 & 59.28\\
NT with data augmentation & 83.42 & 64.85 & 80.15 & 58.65\\
\hline
NT with reads & 82.85 & 61.02 & 68.32 & 57.06\\
NT with reads and data augmentation & 83.11 & 62.82 & 74.81 & 57.32\\
\hline
\end{tabular}

\caption{\label{citation-guide}
Result of data augmentation for the class which has small samples. Data augmentation increases the F1 score.
}
\end{table*}

\subsection{Weighted Soft-voting Ensemble}

To combine the pre-trained nucleotide sequence-based language model and the pre-trained text-based language model mentioned earlier, we use a soft-voting ensemble model. Additionally, we find the optimal weights through validation data and apply them to create a weighted soft voting ensemble model. A more detailed explanation of the validation data will be provided in the Experiment section. This data is a third dataset separate from the training and test data. This allows us to use both nucleotide sequence information and the text information that describes it. This model requires both types of input. It receives the nucleotide sequence and information about Gene Family and Resistance Mechanism in the format [Resistance Mechanism]: Antibiotic Effuls, \#[Gene Family]: Bata-Lactamases\#.

\subsection{Integrating Classes Based on Antibiotic Resistance Ontology}
The databases provided in the literature (CARD, MEGARes) have different classification systems and hierarchical relationships. EBI ARO provides hierarchical information on antibiotic resistance genes. EBI stands for European Bioinformatics Institute. These diverse antibiotic resistance classification systems, gene groupings, and resistance mechanisms can be combined through the EBI ontology, and the model can store integrated concept representations. Each database's header is read and the EBI API is searched. The mapped items are used as new Gene Family. Rather than using very small and specific hierarchical classes, more general hierarchical classes are employed. The third level from the top in the EBI ARO hierarchy is used as the basis.

\begin{figure}[h!]
    \centering
    \includegraphics[width=1\linewidth]{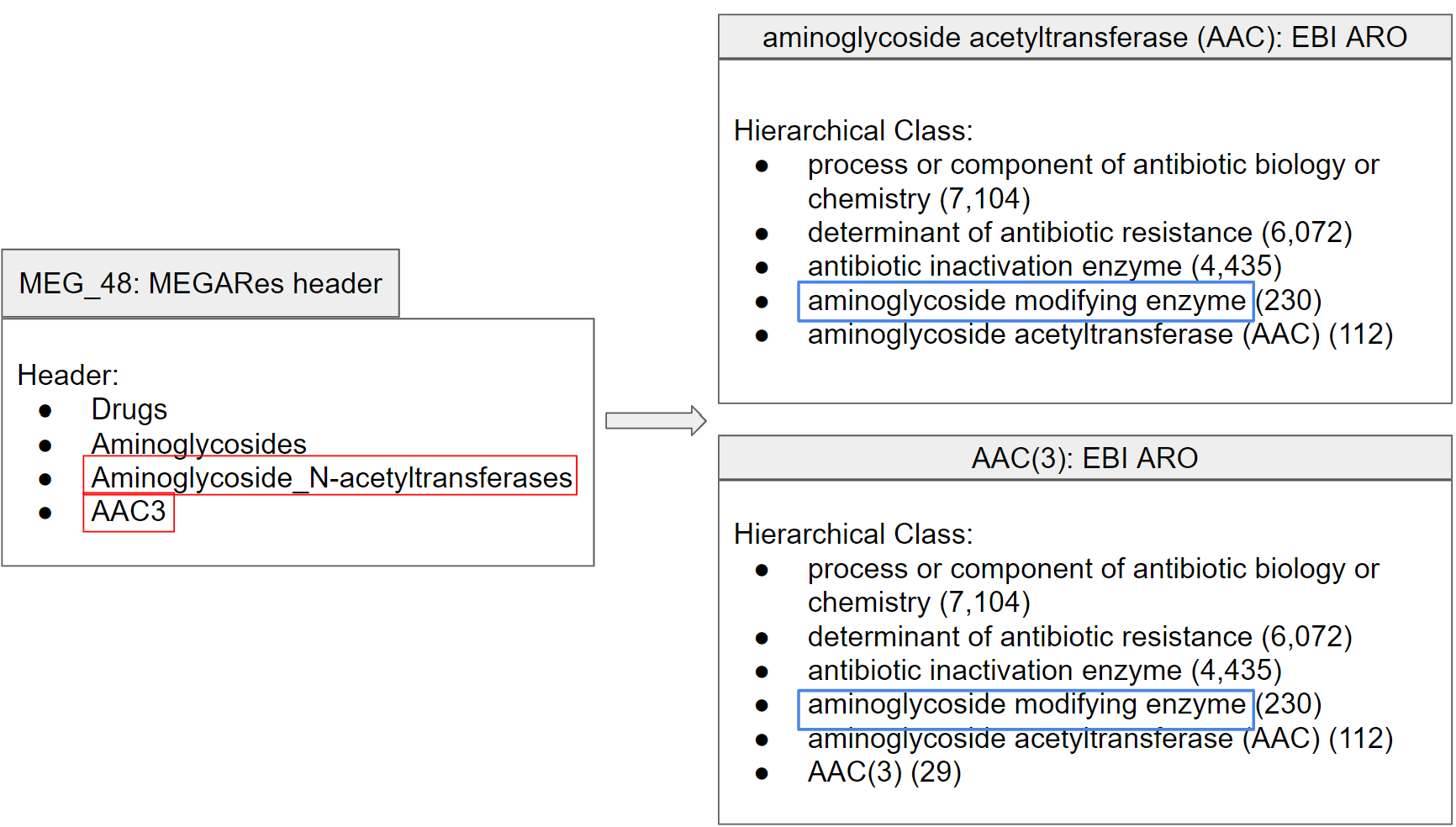}
    \caption{EBI ARO Gene Family mapping: search to find mapping information with header and ontology by using API.}
\end{figure}

\subsection{Data Augmentation Using a Large Language Model}
The categories were integrated based on the EBI ARO Ontology's gene group and CARD Resistance Mechanism. However, there are still cases where the number of samples corresponding to a class is small. Data augmentation was conducted for these cases. BioGPT was used for data augmentation. Similar data were created through prompting. Through this, it was possible to see that performance improved as follows: In particular, the accuracy in classes with a small number of samples increased.

\section{Experiments}

\begin{table*}[h!]
\centering
\begin{tabular}{llllll}
\hline
\textbf{Dataset} &\textbf{Method} & \textbf{Accuracy} & \textbf{Macro F1} & \textbf{Precision} & \textbf{Recall}\\
\hline
CARD & NT & 87.92 &63.08 & 66.46 & 61.51 \\
CARD & BB & 97.22 & 89.68 & 92.09 &90.54 \\
CARD & Ensemble & 97.55 & 93.44 &95.72 &92.86\\
\hline
MEGARes & NT &89.61 &46.42 &54.92 &43.94 \\
MEGARes & BB &99.64 &99.47 &99.96 &99.03\\
MEGARes & Ensemble &99.99 &99.99 &99.99 &99.99\\
\hline
Integrated & NT & 82.89 & 65.79 & 81.84 & 58.67 \\
Integrated & BB & 90.26 & 79.34 & 84.05  & 77.14\\
Integrated & Ensemble &92.11 & 80.95 & 83.52 & 78.94\\
\hline
Integrated with reads & NT & 83.11 & 62.82 & 74.81 & 57.32 \\
Integrated with reads & BB & 90.24 & 79.34 & 84.05  & 77.14\\
Integrated with reads & Ensemble &93.40& 81.85 & 84.34 & 80.25\\
\hline
\end{tabular}
\caption{\label{citation-guide}
Result of using the CARD, MEGARes, and Integrated databases for antibiotic resistance drug class prediction using Nucleotide Transformer(NT), BioBERT(BB), and a weighted ensemble of both. The weighted ensemble with Nucleotide Transformer(NT) and BioBERT(BB) shows better performance in every datasets.
}
\end{table*}

\subsection{Datasets}

The CARD and MEGARes v3 datasets are used for training and evaluation.
Classes with fewer than 15 samples are removed because obtaining meaningful results from the data split is difficult.
The remaining data is split into 75\% for training data, 20\% for test data, and 5\% for validation data.
EBI ARO ontology search is used to integrate the data, which is then split similarly to the above.
Classes with difficult-to-obtain meaningful results are also removed.
The MEGARes dataset consists of 9733 Reference Sequences, 1088 SNPs, 4 antibiotic types, 59 resistance classes, and 233 mechanisms.
The CARD dataset consists of 5194 Reference Sequences and 2005 SNPs, 142 Drug Classes, 331 Gene Families, and 10 Resistance Mechanisms.
The EBI ARO ontology provides hierarchical group information for genes.
Using the EBI ARO Ontology, Gene Family class information can be integrated into a higher-level hierarchy.
The number of Gene Family text information classes in the case of MEGARes is 589, while for CARD, it is 331.
There are 300 and 166 datasets with only one sample in their respective classes for Gene Family in the case of MEGARes and CARD, respectively.
Resistance Mechanism is integrated based on the 6 categories of CARD.
The original 8 categories were reduced to 6, excluding cases of various class combinations and those with very few samples.
Drug Class is integrated using 9 common Drug Classes found in competing models.
Integration is done based on names and theories and has been verified.
Macro f1 score, accuracy, balanced accuracy, and precision are used as performance metrics, and the results are listed in the table 3.

\subsection{Implementation Details}


Basic structure of the model and fine-tuning follow the methods proposed by BioBERT and Nucleotide Transformer.
The layers and information of the model are in the Appendix.

\subsection{Main Results}

Tables 3 show metrics using our method with the latest techniques (SOTA) in the text-based information model for the CARD and MEGARes experiments, showing that our method surpasses previous SOTA.
Additionally, the method using integrated data shows superiority over previous SOTA.
Our method also demonstrates competitive results compared to other competing models and SOTA.\\

\section{Discussion}
\section*{Does text information help?}

In all datasets, using a text information-based language model shows a 9.53 accuracy and 30.34 macro f1 score improvement in CARD and 10.38 accuracy and 50.57 macro f1 score improvement in MEGARes.
Adjusted ratio ensemble models show better performance compared to other cases through experiments.
Existing NT and other nucleotide sequence-based models find it difficult to process natural language.
Our fine-tuned text-based language model was trained using a small amount of pre-training resources (40GB A100 GPU).
By constructing an ensemble model, it achieves better performance compared to competing models such as AMR-meta \cite{marini2022amrmeta}, Meta-MARC \cite{lakin2019hierarchical}, and Deep ARG \cite{arangoargoty2018deeparg}.

\section*{Does text information class integration help?}

To compare with other models, we integrated the class system.
This enables comparison with competing models.
It also allows us to create models for predicting Gene Family and Resistance Mechanism.
In particular, the number of samples corresponding to classes in Gene Family and Resistance Mechanism is very small in many cases.
This integration helps to implement Gene Family and Resistance Mechanism prediction models.
The integrated class system shows better performance compared to cases where it is not.
The number of genes available for training increases.

\section*{Sequencing Read Generation}

In some competing models, it is recommended to use reads instead of full genes.
In the case of AMR-meta, it aims to predict paired end genes.
To compare with these models, it is necessary to generate reads.
Reads generation uses ART.
ART is a simulator for analyzing nucleotide sequences, and it helps with accurate modeling of biological information data as a software \cite{huang2012art}.
ART has the advantage of customizable indel error rates \cite{milhaven2023performance}.
The learning and experiments using these reads are presented in Table .
In this experiment, the proposed model also demonstrates strong competitiveness.\\

\section{Related Work}
\textbf{AMR-meta} is a method for classifying antibiotic resistance in high-speed metagenomic data.
This method uses a sequence alignment-free approach based on k-mers and meta-features, and it utilizes both resistant and non-resistant genes as training data.
As a result, AMR-meta can more accurately identify antibiotic resistance genes and reduce false-positive rates for non-resistant genes.
However, it uses a complex matrix decomposition method to generate meta-features, which can be computationally intensive.
Additionally, the prediction performance of AMR-meta may vary depending on the type of antibiotic used or the diversity of the resistance genes.
These characteristics make AMR-meta useful for analyzing high-speed metagenomic data, but at the same time, they suggest that it may be limited in certain situations.\\
\textbf{AMR++} is a customized bioinformatics pipeline that uses high-throughput sequencing data to predict the diversity and abundance of antibiotic resistance genes (ARGs).
This pipeline is integrated with the MEGARes database, allowing for efficient analysis of ARGs in large-scale metagenomic sequencing data.
The main advantage of AMR++ is its high throughput and efficiency, enabling users to quickly and accurately analyze complex datasets.
In addition, this software can distinguish between types of ARGs, including cases where resistance genes require specific mutations.
However, this pipeline requires high-quality assembled and/or translated data, which may cause difficulties or limitations in generating metagenomic datasets.
Furthermore, AMR++ may require advanced bioinformatics skills and resources, potentially limiting accessibility for some researchers.\\
\textbf{Meta-MARC} is a machine learning classifier developed to enhance the detection and classification of antibiotic resistance genes.
This system is based on the MEGARes database and uses DNA-based hierarchical Hidden Markov Models (HMMs) to classify antibiotic resistance genes in high-throughput sequencing data.
Meta-MARC is robust against various gene mutations, which is particularly useful for non-standard databases and sequences.
This tool provides high sensitivity and specificity, playing a crucial role in accurate antibiotic resistance detection.
However, Meta-MARC is computationally demanding, particularly when dealing with large datasets, which can result in increased processing time and memory usage.
Additionally, high sensitivity settings may potentially increase false positives, so users must carefully interpret the results.\\
\textbf{DeepARG} is a deep learning-based system used for predicting antibiotic resistance genes (ARGs) in metagenomic data.
It utilizes two models, DeepARG-SS and DeepARG-LS, for classifying short and full-length gene sequences.
Compared to the traditional 'best hit' approach, it has the advantage of identifying a wider range of ARG diversity with lower false negative rates.
However, the performance of this system heavily depends on the quality of the training database, and it has limitations when it comes to predicting new categories of ARGs.
Despite these limitations, DeepARG is a useful tool for evaluating the presence and diversity of ARGs in environmental samples.\\
\section{Conclusion}
As far as we know, our work is the first to combine natural language models and biological sequence models to predict antibiotic resistance genes.
We proposed a model that combines two different attribute language models into an ensemble.
By using both nucleotide sequence information and its description, including Gene family and resistance mechanism information, it enables more accurate drug class predictions.
We also integrated various databases using the EBI ontology and used a large language model (LLM) for data augmentation in classes with insufficient data.
As a result, we achieved performance close to the state-of-the-art.
We believe this fusion has significant meaning.
Moreover, we tested the structure we trained using only nucleotide sequences and obtained acceptable results.
This seems promising for future research.










\section*{Acknowledgements}
This work is supported by the National Science Foundation.

\bibliography{anthology,custom}

\begin{thebibliography}{21}
\expandafter\ifx\csname natexlab\endcsname\relax\def\natexlab#1{#1}\fi

\bibitem[{Arango-Argoty et~al.(2018)Arango-Argoty, Garner, Pruden, Heath, Vikesland, and Zhang}]{arangoargoty2018deeparg}
Gustavo Arango-Argoty, Emily Garner, Amy Pruden, Lenwood~S. Heath, Peter Vikesland, and Liqing Zhang. 2018.
\newblock \href {https://doi.org/10.1186/s40168-018-0401-z} {Deeparg: A deep learning approach for predicting antibiotic resistance genes from metagenomic data}.
\newblock \emph{Microbiome}, 6(1):23.

\bibitem[{Bonin et~al.(2023)Bonin, Doster, Worley, Pinnell, Bravo, Ferm, Marini, Prosperi, Noyes, Morley, and Boucher}]{Bonin2023MEGAResAMR}
Nathalie Bonin, Enrique Doster, Hannah Worley, Lee~J Pinnell, Jonathan~E Bravo, Peter Ferm, Simone Marini, Mattia Prosperi, Noelle Noyes, Paul~S Morley, and Christina Boucher. 2023.
\newblock \href {https://doi.org/10.1093/nar/gkac1047} {Megares and amr++, v3.0: an updated comprehensive database of antimicrobial resistance determinants and an improved software pipeline for classification using high-throughput sequencing}.
\newblock \emph{Nucleic Acids Research}, 51(D1):D744--D752.

\bibitem[{Brandes et~al.(2022)Brandes, Ofer, Peleg, Rappoport, and Linial}]{Brandes2022ProteinBERT}
Nadav Brandes, Dan Ofer, Yam Peleg, Nadav Rappoport, and Michal Linial. 2022.
\newblock \href {https://doi.org/10.1093/bioinformatics/btac020} {Proteinbert: a universal deep-learning model of protein sequence and function}.
\newblock \emph{Bioinformatics}, 38(8):2102--2110.

\bibitem[{Cook et~al.(2016)Cook, Bergman, Finn, Cochrane, Birney, and Apweiler}]{cook2016european}
Charles~E. Cook, Mary~Todd Bergman, Robert~D. Finn, Guy Cochrane, Ewan Birney, and Rolf Apweiler. 2016.
\newblock \href {https://doi.org/10.1093/nar/gkv1352} {The european bioinformatics institute in 2016: Data growth and integration}.
\newblock \emph{Nucleic Acids Research}, 44(D1):D20--D26.

\bibitem[{Dalla-Torre et~al.(2023)Dalla-Torre, Gonzalez, Mendoza-Revilla, Carranza, Grzywaczewski, Oteri, Dallago et~al.}]{dallatorre2023nucleotide}
Hugo Dalla-Torre, Liam Gonzalez, Javier Mendoza-Revilla, Nicolas~Lopez Carranza, Adam~Henryk Grzywaczewski, Francesco Oteri, Christian Dallago, et~al. 2023.
\newblock \href {https://doi.org/10.1101/2023.01.11.523679} {The nucleotide transformer: Building and evaluating robust foundation models for human genomics}.
\newblock \emph{Genomics}.

\bibitem[{Doster et~al.(2020)Doster, Lakin, Dean, Wolfe, Young, Boucher, Belk, Noyes, and Morley}]{Doster2020}
Enrique Doster, Steven~M Lakin, Christopher~J Dean, Cory Wolfe, Jared~G Young, Christina Boucher, Keith~E Belk, Noelle~R Noyes, and Paul~S Morley. 2020.
\newblock \href {https://doi.org/10.1093/nar/gkz1010} {Megares 2.0: a database for classification of antimicrobial drug, biocide and metal resistance determinants in metagenomic sequence data}.
\newblock \emph{Nucleic Acids Research}, 48(D1):D561--D569.

\bibitem[{Hu et~al.(2021)Hu, Shen, Wallis, Allen-Zhu, Li, Wang, Wang, and Chen}]{hu2021lora}
Edward~J. Hu, Yelong Shen, Phillip Wallis, Zeyuan Allen-Zhu, Yuanzhi Li, Shean Wang, Lu~Wang, and Weizhu Chen. 2021.
\newblock \href {https://arxiv.org/abs/2106.09685v2} {Lora: Low-rank adaptation of large language models}.
\newblock \emph{arXiv}.
\newblock ArXiv:2106.09685v2.

\bibitem[{Huang et~al.(2012)Huang, Li, Myers, and Marth}]{huang2012art}
Weichun Huang, Leping Li, Jason~R. Myers, and Gabor~T. Marth. 2012.
\newblock \href {https://doi.org/10.1093/bioinformatics/btr708} {Art: A next-generation sequencing read simulator}.
\newblock \emph{Bioinformatics}, 28(4):593--594.

\bibitem[{Ji et~al.(2021)Ji, Zhou, Liu, and Davuluri}]{Ji2021DNABERT}
Yanrong Ji, Zhihan Zhou, Han Liu, and Ramana~V Davuluri. 2021.
\newblock \href {https://doi.org/10.1093/bioinformatics/btab083} {Dnabert: pre-trained bidirectional encoder representations from transformers model for dna-language in genome}.
\newblock \emph{Bioinformatics}, 37(15):2112--2120.

\bibitem[{Jia et~al.(2017)Jia, Raphenya, Alcock, Waglechner, Guo, Tsang, Lago, Dave, Pereira, Sharma, Doshi, Courtot, Lo, Williams, Frye, Elsayegh, Sardar, Westman, Pawlowski, Johnson, Brinkman, Wright, and McArthur}]{Jia2017CARD}
Baofeng Jia, Amogelang~R. Raphenya, Brian Alcock, Nicholas Waglechner, Peiyao Guo, Kara~K. Tsang, Briony~A. Lago, Biren~M. Dave, Sheldon Pereira, Arjun~N. Sharma, Sachin Doshi, Mélanie Courtot, Raymond Lo, Laura~E. Williams, Jonathan~G. Frye, Tariq Elsayegh, Daim Sardar, Erin~L. Westman, Andrew~C. Pawlowski, Timothy~A. Johnson, Fiona~S.L. Brinkman, Gerard~D. Wright, and Andrew~G. McArthur. 2017.
\newblock \href {https://doi.org/10.1093/nar/gkw1004} {Card 2017: expansion and model-centric curation of the comprehensive antibiotic resistance database}.
\newblock \emph{Nucleic Acids Research}, 45(D1):D566--D573.

\bibitem[{Kang et~al.(2022)Kang, Goo, Lee, Chae, Yun, and Jung}]{Kang2022FineTuningBERT}
Hyeunseok Kang, Sungwoo Goo, Hyunjung Lee, Jung-Woo Chae, Hwi-Yeol Yun, and Sangkeun Jung. 2022.
\newblock \href {https://doi.org/10.3390/pharmaceutics14081710} {Fine-tuning of bert model to accurately predict drug-target interactions}.
\newblock \emph{Pharmaceutics}, 14(8):1710.

\bibitem[{Kumari et~al.(2021)Kumari, Kumar, and Mittal}]{Kumari2021}
Saloni Kumari, Deepika Kumar, and Mamta Mittal. 2021.
\newblock An ensemble approach for classification and prediction of diabetes mellitus using soft voting classifier.
\newblock \emph{International Journal of Cognitive Computing in Engineering}, 2:40--46.

\bibitem[{Lakin et~al.(2019)Lakin, Kuhnle, Alipanahi, Noyes, Dean, Muggli, Raymond et~al.}]{lakin2019hierarchical}
Steven~M. Lakin, Alan Kuhnle, Bahar Alipanahi, Noelle~R. Noyes, Chris Dean, Martin Muggli, Rob Raymond, et~al. 2019.
\newblock \href {https://doi.org/10.1038/s42003-019-0545-9} {Hierarchical hidden markov models enable accurate and diverse detection of antimicrobial resistance sequences}.
\newblock \emph{Communications Biology}, 2(1):294.

\bibitem[{Lee et~al.(2020)Lee, Yoon, Kim, Kim, Kim, So, and Kang}]{lee2020biobert}
Jinhyuk Lee, Wonjin Yoon, Sungdong Kim, Donghyeon Kim, Sunkyu Kim, Chan~Ho So, and Jaewoo Kang. 2020.
\newblock \href {https://doi.org/10.1093/bioinformatics/btz682} {Biobert: A pre-trained biomedical language representation model for biomedical text mining}.
\newblock \emph{Bioinformatics}, 36(4):1234--1240.

\bibitem[{Luo et~al.(2022)Luo, Sun, Xia, Qin, Zhang, Poon, and Liu}]{luo2022biogpt}
Renqian Luo, Liai Sun, Yingce Xia, Tao Qin, Sheng Zhang, Hoifung Poon, and Tie-Yan Liu. 2022.
\newblock \href {https://doi.org/10.1093/bib/bbac409} {Biogpt: Generative pre-trained transformer for biomedical text generation and mining}.
\newblock \emph{Briefings in Bioinformatics}, 23(6):bbac409.

\bibitem[{Marini et~al.(2022)Marini, Oliva, Slizovskiy, Das, Noyes, Kahveci, Boucher, and Prosperi}]{marini2022amrmeta}
Simone Marini, Marco Oliva, Ilya~B Slizovskiy, Rishabh~A Das, Noelle~Robertson Noyes, Tamer Kahveci, Christina Boucher, and Mattia Prosperi. 2022.
\newblock \href {https://doi.org/10.1093/gigascience/giac029} {Amr-meta: A k -mer and metafeature approach to classify antimicrobial resistance from high-throughput short-read metagenomics data}.
\newblock \emph{GigaScience}, 11.
\newblock Giac029.

\bibitem[{Mej{\'i}a-Guerra and Buckler(2019)}]{mejiaguerra2019kmer}
Mar{\'i}a~Katherine Mej{\'i}a-Guerra and Edward~S. Buckler. 2019.
\newblock \href {https://doi.org/10.1186/s12870-019-1693-2} {A k-mer grammar analysis to uncover maize regulatory architecture}.
\newblock \emph{BMC Plant Biology}, 19(1):103.

\bibitem[{Milhaven and Pfeifer(2023)}]{milhaven2023performance}
Mark Milhaven and Susanne~P. Pfeifer. 2023.
\newblock \href {https://doi.org/10.1038/s41437-022-00577-3} {Performance evaluation of six popular short-read simulators}.
\newblock \emph{Heredity}, 130(2):55--63.

\bibitem[{Zhang et~al.(2022)Zhang, Zhang, Wang, Xu, Lu, Hong, Penuelas, Gillings, Wang, Gao, and Qian}]{Zhang2022GlobalHealthRisk:1}
Zhenyan Zhang, Qi~Zhang, Tingzhang Wang, Nuohan Xu, Tao Lu, Wenjie Hong, Josep Penuelas, Michael Gillings, Meixia Wang, Wenwen Gao, and Haifeng Qian. 2022.
\newblock Assessment of global health risk of antibiotic resistance genes.
\newblock \emph{Nature Communications}, 13.

\bibitem[{Zhou and Chen(2021)}]{zhou2021improved}
Wenxuan Zhou and Muhao Chen. 2021.
\newblock \href {https://arxiv.org/abs/2102.01373v4} {An improved baseline for sentence-level relation extraction}.
\newblock \emph{arXiv}.
\newblock ArXiv:2102.01373v4.

\bibitem[{Zhou et~al.(2023)Zhou, Ji, Li, Dutta, Davuluri, and Liu}]{zhou2023dnabert2}
Zhihan Zhou, Yanrong Ji, Weijian Li, Pratik Dutta, Ramana Davuluri, and Han Liu. 2023.
\newblock \href {https://arxiv.org/abs/2306.15006v1} {Dnabert-2: Efficient foundation model and benchmark for multi-species genome}.
\newblock \emph{arXiv}.
\newblock ArXiv:2306.15006v1.

\end{thebibliography}




\end{document}